%% file: main.tex
\documentclass{article}
\PassOptionsToPackage{numbers,square,sort&compress}{natbib}
\usepackage[preprint]{neurips_2025}
\usepackage[T1]{fontenc}
\usepackage{newtxtext,newtxmath}
\usepackage{microtype}
\usepackage{amsmath}
\usepackage{booktabs}
\usepackage{xcolor}
\usepackage{graphicx}
\usepackage{caption}
\usepackage{subcaption}
\usepackage{enumitem}
\usepackage{pifont}
\usepackage{algorithm}
\usepackage[noend]{algpseudocode}
\usepackage{tikz}
\usetikzlibrary{arrows.meta,positioning,calc}
\usepackage{pgfplots}
\pgfplotsset{compat=1.18}
\usepgfplotslibrary{groupplots}
\usepackage[hidelinks]{hyperref}
\usepackage[capitalise,noabbrev]{cleveref}

\newcommand{\sys}{DEVL}


\newcommand{\ystar}{y^{\star}}
\newcommand{\G}{\mathcal{G}}

\definecolor{spec}{HTML}{1f6feb}
\definecolor{canon}{HTML}{8c8c8c}
\definecolor{official}{HTML}{bf6a02}
\definecolor{sglang}{HTML}{0e8a7d}
\definecolor{vllm}{HTML}{8250df}
\definecolor{accept}{HTML}{2da44e}
\definecolor{reject}{HTML}{cf222e}

\title{Decode Early, Verify Late:\\Streaming Offline Speech Recognizers\\Without Accuracy Loss}
\author{Doyeob (Toby) Kim\\Nari Labs\\\texttt{toby@narilabs.com}}

\begin{document}
\maketitle

\begin{abstract}
Voice agents need accurate transcripts soon after a user stops speaking. We propose Decode Early, Verify Late (\sys{}), a training-free framework for streaming inference with offline ASR models. The server keeps partial transcripts as drafts and verifies the latest against the complete audio in one forward pass at turn end. Under stated input and decoding conditions, the final equals the model's greedy offline output in exact arithmetic. We adapt \sys{} to four model families. In our five-recognizer comparison, 95--100\% of final transcripts are byte-identical to the offline transcripts, WER stays within 0.13 points of offline, and single-stream time to final falls 1.5--7.3$\times$. Serving Qwen3-ASR-1.7B on one A100, \sys{} returns finals a median 22--26 ms after turn end at up to 64 streams, 2.7--3.9$\times$ sooner than the official streaming mode. It is the only system we tested that meets a 100 ms p99 target, up to 80 streams on Pipecat-STT and 64 on WildASR.
\end{abstract}

\section{Introduction}
\label{sec:intro}

A streaming recognizer shows \emph{partials} during speech and returns a \emph{final} transcript when the turn ends, at the \emph{commit}. The \emph{time to final segment} (TTFS), from commit to final, adds to the delay before a voice agent answers. A GPU's \emph{capacity} is the number of streams it serves within a TTFS target.

The most accurate open-weight recognizers are \emph{offline} models, which decode a complete utterance at once. On the Open ASR Leaderboard~\cite{srivastav2025openasr} (10 November 2026), the top three open-weight entries are all offline: a Qwen3-ASR fine-tune, NVIDIA-Nemotron-3.5-Transcribe-30B-A4B~\cite{nemotrontranscribe2026} and Qwen3-ASR-1.7B~\cite{qwen3asr2026}. The best streaming entry ranks 38th of 64.

Existing streaming options cost accuracy or time. Qwen3-ASR's toolkit fixes earlier output as a prefix~\cite{qwen3asrtoolkit}; NVIDIA Nemotron and Microsoft VibeVoice offer separate streaming models~\cite{nemotronstreaming2026,vibevoicestream2026}; Whisper-Streaming keeps text that consecutive hypotheses agree on~\cite{machacek2023whisperstreaming,liu2020partial}. Each has higher WER than offline decoding in our engine (\cref{tab:models}, with a decoding-settings caveat for Whisper). vLLM's realtime mode for Qwen3-ASR restates earlier segments and scores 47.7\% (Pipecat-STT) and 54.2\% (WildASR) WER (\cref{sec:eval}). Alternatively, vLLM~\cite{kwon2023vllm} and SGLang-Omni~\cite{sglangomni2026} decode the complete utterance at the commit, placing the whole decode between the user's last word and the agent's reply.

Decode Early, Verify Late (\sys{}) reuses the partials a streaming server already computes. With Qwen3-ASR, the last partial matches the first 88\% of the final's words in the median on Pipecat-STT~\cite{pipecatstt} and 82\% on WildASR~\cite{wildasr}. \sys{} keeps each partial as a \emph{draft}. At the commit, one forward pass checks it against the complete audio, as in speculative decoding~\cite{leviathan2023specdec}, and decoding resumes after the longest correct prefix.

We state conditions for exact greedy offline finals (\cref{sec:design}), adapt \sys{} to four model families (\cref{tab:arch}), and evaluate a scheduler that prioritizes finals over drafts on five recognizers, two datasets and 1--128 concurrent streams (\cref{sec:eval}).

Developed in summer 2026, \sys{} powers our hosted Qwen3-ASR service, which ranked first in TTFS (44 ms median) and second in WER (3.6\%) on Coval's speech-to-text benchmark in September 2026~\cite{narilabscoval2026}.

\section{Method}
\label{sec:design}

An offline recognizer encodes audio $x$ into embeddings that form part of the text decoder's prompt. The decoder \emph{prefills} the prompt, caches its state, then generates tokens greedily until end-of-sequence. We write this output as $\G(x)$.

\begin{figure}[t]
\centering
\begin{tikzpicture}[x=0.93cm,y=0.55cm,font=\footnotesize,>={Stealth[length=4pt]}]
  \fill[blue!15] (0,0.3) rectangle (8,1.0);
  \node at (4,0.65) {audio blocks: encoded and prefilled while streaming};
  \fill[orange!25] (8,0.3) rectangle (10.4,1.0);
  \node at (9.2,0.65) {tail};
  \draw[->] (0,0) -- (11.2,0) node[right] {audio (s)};
  \foreach \t in {0,2,4,6,8,10} \draw (\t,0.1) -- (\t,-0.1) node[below] {\t};
  \foreach \t/\k in {1.8/1,3.6/2,5.4/3,7.2/4,9.0/5} {
    \draw[densely dotted] (\t,1.0) -- (\t,1.7);
    \fill[spec] (\t,1.7) circle (2pt);
    \node[above,spec] at (\t,1.75) {$d^{(\k)}$};
  }
  \node[anchor=west,spec] at (-0.1,3.0) {decode early: checkpoints decode the audio so far, publish a partial, keep it as a draft};
  \draw[very thick,reject] (10.4,-0.2) -- (10.4,2.6);
  \node[reject,anchor=west] at (10.45,2.35) {commit};
  \draw[->,thick] (9.12,1.75) to[bend left=40] (10.35,1.3);
  \node[anchor=west] at (10.45,1.45) {verify late};
\end{tikzpicture}
\caption{\textbf{One stream in \sys{}.} Audio is encoded and prefilled into the KV cache as it arrives. Checkpoints decode the audio so far and keep each result as a draft $d^{(k)}$. At the commit, the server prefills only the tail, verifies the last draft in the same pass, and decodes only after its longest correct prefix.}
\label{fig:timeline}
\end{figure}
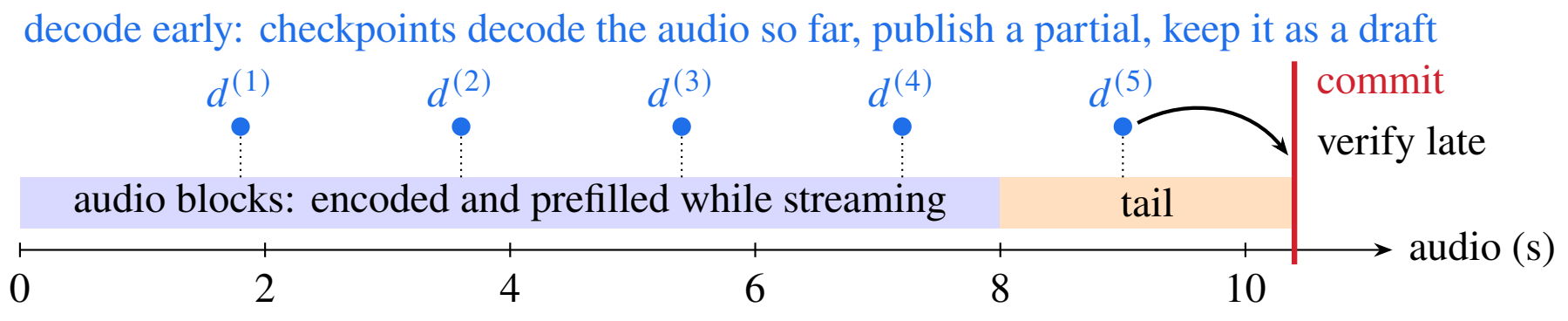

\sys{} targets $\G(x)$ with the server's own weights and kernels. It requires greedy decoding, a deterministic front end and decoder state that can be restored to an accepted prefix. Whenever a draft is checked, the audio representations, prompt, token positions and retained decoder state must equal those of an offline pass over the same audio. Cached representations are recomputed if new audio changes them. Four design choices maintain this invariant for Qwen3-ASR (\cref{fig:timeline,alg:devl}).

First, \sys{} encodes fixed audio blocks counted from the stream's start, independent of the client's chunk boundaries. Each valid block is prefilled into the KV cache once. If input normalization changes, all completed blocks and their decoder state are rebuilt. At the commit, only the \emph{tail} after the last cached block remains.

Second, \emph{checkpoints} produce partials during speech. A checkpoint copies the stream's KV cache, prefills uncached audio and the prompt suffix, then checks the previous draft by the same rule as the final. It decodes beyond the accepted prefix, sends the result as a partial and keeps it as the next draft, recording the audio it saw.

Third, \sys{} checks the latest draft in the same pass that prefills the audio tail and prompt suffix. Up to the first mismatch, the draft equals greedy decoding~\cite{leviathan2023specdec}. The server rolls back to that prefix and continues decoding. It can realign the remaining draft after a correction and verify it again. A completed checkpoint that saw all the audio can be promoted directly to the final. If no draft is usable, the server decodes normally.

Fourth, the scheduler runs pending finals before checkpoints. A checkpoint delays a final by at most the GPU step already running; finals can still queue behind other finals. Checkpoints use remaining GPU time, with at most four streams active except when epoch cuts require more. The engine uses a paged KV cache~\cite{kwon2023vllm} and CUDA graphs.

Under these conditions, finals equal $\G(x)$ in exact arithmetic (proof in \cref{app:exact}). Our Qwen3-ASR engine also cuts utterances longer than 10 s into \emph{epochs} at sentence ends to shorten the final decode. Epoch-cut finals are outside this guarantee.

\label{sec:other-design}
\textbf{Other architectures.} \Cref{tab:arch} adapts these choices to Whisper~\cite{radford2023whisper}, VibeVoice-ASR~\cite{vibevoiceasr2026} and NVIDIA-Nemotron-3.5-Transcribe~\cite{nemotrontranscribe2026}. Windowed or causal encoders can reuse audio representations; whole-clip encoders re-encode all audio at the commit. For VibeVoice-ASR, we remove input gain and use the mean acoustic latent to make the front end deterministic. Nemotron restores its recurrent Mamba-2 state by replaying accepted inputs, as in ReplaySSM~\cite{replayssm2026}.

\begin{table}[t]
\centering\footnotesize
\caption{\textbf{How \sys{} adapts to each model family.} Cached in speech: work reused at the commit. Commit work: what is left when a draft is available. Nemotron replays its Mamba-2 state. Only the Qwen3-ASR engine cuts long utterances into epochs.}
\label{tab:arch}
\setlength{\tabcolsep}{4pt}
\begin{tabular}{@{}lllll@{}}
\toprule
model & encoder sees & cached in speech & commit work & rollback \\
\midrule
Qwen3-ASR & fixed windows & encoder blocks, decoder KV & prefill tail and draft & truncate KV \\
Whisper & whole clip & -- & encode all; prefill and draft & overwrite KV \\
VibeVoice-ASR & causal chunks & encoder output & encode tail; prefill and draft & overwrite KV \\
Nemotron & whole clip & -- & encode all; prefill and draft & replay state \\
\bottomrule
\end{tabular}
\end{table}

\section{Evaluating Streaming Recognizers for Voice Agents}
\label{sec:eval}

We measure TTFS under load and compare final transcripts with each model's offline decode in our engine.

\subsection{Data and Benchmarks}
\label{sec:setup}

\textbf{Pipecat-STT.} The 897 voice-agent turns of Pipecat's stt-benchmark-data~\cite{pipecatstt}, as packaged in Coval's stt-v3 set~\cite{sttv3}. Its reference transcripts were generated by Gemini, not by humans.

\textbf{WildASR.} Coval's seven WildASR sets~\cite{wildasr,sttv3}, with 2,549 English clips. Six sets play FLEURS sentences under different recording conditions, and one holds accented speech.

\textbf{Load and metrics.} A closed loop keeps $C$ streams open, each starting its next utterance when the previous final arrives. Streams send 100 ms chunks in real time and commit with the last chunk. TTFS is measured at the client, with 95\% bootstrap intervals. Capacity is the largest measured load meeting the target at that and every lower load. \emph{Identical} compares final transcript text byte for byte with the offline transcript. For VibeVoice-ASR, this is the joined JSON \texttt{Content} text, excluding timestamps and speaker metadata.

\subsection{Baselines}

Our engine serves Qwen3-ASR-1.7B on one A100-SXM4-40GB. The \emph{no-drafts server} is the same engine with checkpoints and epochs off. The \emph{offline reference} decodes each complete utterance alone on it. We compare against Qwen3-ASR's toolkit (qwen-asr 0.0.6)~\cite{qwen3asrtoolkit} in its \emph{official streaming mode}. Every 2 s, it re-decodes all audio so far, with its earlier output as a fixed prefix. We also compare against SGLang-Omni 0.2.0 and vLLM 0.33.0, which transcribe the complete utterance at the commit. We exclude vLLM 0.33.0's realtime mode, which restates earlier 5 s segments in 48\% (Pipecat-STT) and 58\% (WildASR) of utterances; vLLM-Omni does not serve Qwen3-ASR as a standalone recognizer.

We also run Whisper large-v3-turbo and large-v3, VibeVoice-ASR and NVIDIA-Nemotron-3.5-Transcribe-30B-A4B~\cite{nemotrontranscribe2026}. Each is compared with its own offline decode and with its usual streaming option.

\section{Results}
\label{sec:results}

\subsection{Accuracy across five recognizers}
\label{sec:eval-accuracy}
\label{sec:eval-exact}

\begin{table}[t]
\centering\small
\caption{\textbf{Accuracy and TTFS of five recognizers.} Identical: byte-identical final transcript text (\cref{sec:setup}). TTFS: median at one stream, drafts off $\to$ on. Qwen3-ASR uses client timing with epochs off; its WER and identical share come from the 64-stream run. Other models use server timing. Streaming: each model's usual option; $^\S$Whisper-Streaming~\cite{machacek2023whisperstreaming} on another backend, with most of its gap due to default decoding settings; $^\ddagger$Microsoft's separate VibeVoice-ASR-Streaming-7B~\cite{vibevoicestream2026}; $^\dagger$NVIDIA's separate nemotron-3.5-asr-streaming-0.6b~\cite{nemotronstreaming2026}. Samples in \cref{app:models}.}
\label{tab:models}
\setlength{\tabcolsep}{3.5pt}
\begin{tabular}{@{}llccccc@{}}
\toprule
& & \multicolumn{3}{c}{WER (\%)} & & TTFS (ms), \\
\cmidrule(lr){3-5}
model & data & offline & streaming & \sys{} & identical & drafts off $\to$ on \\
\midrule
Qwen3-ASR-1.7B & Pipecat-STT  & 1.78 & 1.94 & 1.77 & 98.0\% & 115 $\to$ 24 \\
               & WildASR & 4.75 & 5.14 & 4.65 & 98.5\% & 78 $\to$ 25 \\
Whisper large-v3-turbo & Pipecat-STT  & 3.45 & 10.72$^\S$ & 3.45 & 99.7\% & 58 $\to$ 32 \\
                 & WildASR & 5.70 & 11.76$^\S$ & 5.70 & 100\% & 48 $\to$ 32 \\
Whisper large-v3 & Pipecat-STT  & 3.58 & 6.12$^\S$ & 3.58 & 99.7\% & 223 $\to$ 58 \\
                 & WildASR & 5.05 & 5.54$^\S$ & 5.05 & 99.7\% & 159 $\to$ 59 \\
VibeVoice-ASR & Pipecat-STT  & 2.62 & 4.08$^\ddagger$ & 2.65 & 95.4\% & 888 $\to$ 121 \\
                 & WildASR & 12.44 & 16.92$^\ddagger$ & 12.31 & 95.5\% & 701 $\to$ 135 \\
\begin{tabular}[t]{@{}l@{}}NVIDIA-Nemotron-3.5-\\Transcribe-30B-A4B\end{tabular} & Pipecat-STT  & 1.74 & 5.62$^\dagger$ & 1.72 & 95.0\% & 427 $\to$ 176 \\
                 & WildASR & 4.68 & 11.51$^\dagger$ & 4.63 & 95.1\% & 308 $\to$ 180 \\
\bottomrule
\end{tabular}
\end{table}

In the five-recognizer comparison (\cref{tab:models}), \sys{}'s WER is within 0.13 points of offline decoding, 95--100\% of final transcripts are byte-identical, and single-stream finals arrive 1.5--7.3$\times$ sooner. Whisper-Streaming's accuracy gap mostly comes from its default decoding settings (\cref{app:models}). Where we recorded logits for \sys{}'s differing finals, the first divergence was at a near-tie between the top two tokens. Even without drafts, batching changes up to 1.9\% of Qwen3-ASR's finals at 128 streams (\cref{app:exact}).

\subsection{Qwen3-ASR under load}
\label{sec:eval-latency}

\begin{figure}[t]
\centering
\begin{tikzpicture}
\begin{groupplot}[group style={group size=2 by 1,horizontal sep=1.3cm},
  width=0.5\linewidth,height=3.7cm,xmode=log,log basis x=2,
  xtick={1,8,32,64,128},xticklabels={1,8,32,64,128},
  xlabel={concurrent streams},ymin=0,ymax=250,grid=major,grid style={gray!20},
  legend style={font=\scriptsize,draw=none,fill=none,legend columns=5,/tikz/every even column/.append style={column sep=6pt}},
  tick label style={font=\footnotesize},label style={font=\small}]
\nextgroupplot[title={Pipecat-STT},ylabel={median TTFS (ms)}]
  \addplot[canon,mark=o,thick,error bars/.cd,y dir=both,y explicit] table[x=c,y=p50,y error plus expr=\thisrow{hi}-\thisrow{p50},y error minus expr=\thisrow{p50}-\thisrow{lo}] {data-cap/ttf-stt-v3-N.dat};
  \addplot[spec,mark=*,thick,error bars/.cd,y dir=both,y explicit] table[x=c,y=p50,y error plus expr=\thisrow{hi}-\thisrow{p50},y error minus expr=\thisrow{p50}-\thisrow{lo}] {data-cap/ttf-stt-v3-S.dat};
  \addplot[official,mark=triangle*,thick,dashed,error bars/.cd,y dir=both,y explicit] table[x=c,y=p50,y error plus expr=\thisrow{hi}-\thisrow{p50},y error minus expr=\thisrow{p50}-\thisrow{lo}] {data-cap/ttf-stt-v3-O.dat};
  \addplot[sglang,mark=square*,thick,densely dashed,error bars/.cd,y dir=both,y explicit] table[x=c,y=p50,y error plus expr=\thisrow{hi}-\thisrow{p50},y error minus expr=\thisrow{p50}-\thisrow{lo}] {data-cap/ttf-stt-v3-G.dat};
  \addplot[vllm,mark=diamond*,thick,densely dashed,error bars/.cd,y dir=both,y explicit] table[x=c,y=p50,y error plus expr=\thisrow{hi}-\thisrow{p50},y error minus expr=\thisrow{p50}-\thisrow{lo}] {data-cap/ttf-stt-v3-L.dat};
  \legend{no drafts,\sys{},official streaming,SGLang-Omni,vLLM}
  \pgfplotsset{legend to name=loadleg}
\nextgroupplot[title={WildASR}]
  \addplot[canon,mark=o,thick,error bars/.cd,y dir=both,y explicit] table[x=c,y=p50,y error plus expr=\thisrow{hi}-\thisrow{p50},y error minus expr=\thisrow{p50}-\thisrow{lo}] {data-cap/ttf-wildasr-N.dat};
  \addplot[spec,mark=*,thick,error bars/.cd,y dir=both,y explicit] table[x=c,y=p50,y error plus expr=\thisrow{hi}-\thisrow{p50},y error minus expr=\thisrow{p50}-\thisrow{lo}] {data-cap/ttf-wildasr-S.dat};
  \addplot[official,mark=triangle*,thick,dashed,error bars/.cd,y dir=both,y explicit] table[x=c,y=p50,y error plus expr=\thisrow{hi}-\thisrow{p50},y error minus expr=\thisrow{p50}-\thisrow{lo}] {data-cap/ttf-wildasr-O.dat};
  \addplot[sglang,mark=square*,thick,densely dashed,error bars/.cd,y dir=both,y explicit] table[x=c,y=p50,y error plus expr=\thisrow{hi}-\thisrow{p50},y error minus expr=\thisrow{p50}-\thisrow{lo}] {data-cap/ttf-wildasr-G.dat};
  \addplot[vllm,mark=diamond*,thick,densely dashed,error bars/.cd,y dir=both,y explicit] table[x=c,y=p50,y error plus expr=\thisrow{hi}-\thisrow{p50},y error minus expr=\thisrow{p50}-\thisrow{lo}] {data-cap/ttf-wildasr-L.dat};
\end{groupplot}
\end{tikzpicture}

\ref{loadleg}
\caption{\textbf{Qwen3-ASR TTFS under load, one A100.} Median TTFS with 95\% intervals against concurrent streams, log scale (lower is better). SGLang-Omni and vLLM transcribe the complete utterance at the commit.}
\label{fig:load}
\end{figure}
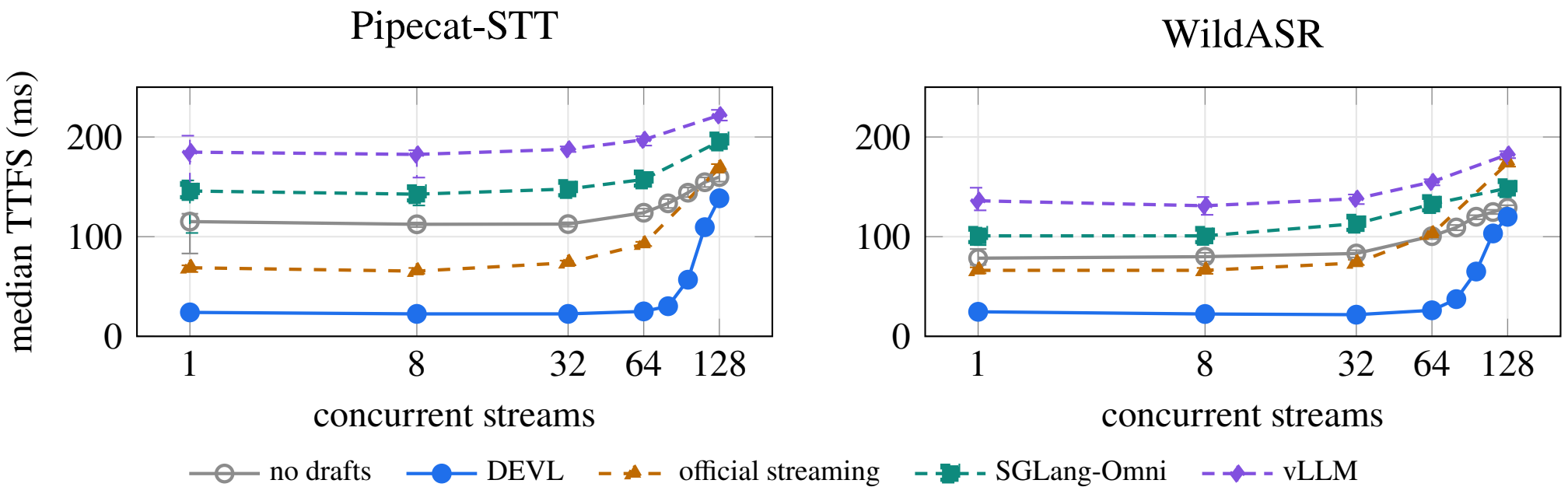

Up to 64 streams, \sys{} returns finals a median 22--26 ms after the commit on both datasets (\cref{fig:load}). That is 2.7--3.9$\times$ sooner than the official streaming mode, which decodes only past its fixed prefix and pays the accuracy cost in \cref{tab:models}. It is 3.2--5.0$\times$ sooner than the no-drafts server, and 4.1--8.3$\times$ sooner than SGLang-Omni and vLLM. Only \sys{} meets a 100 ms p99 target: up to 80 streams on Pipecat-STT and 64 on WildASR. Every other system misses it even at one stream (\cref{tab:capacity}). At a 300 ms p99 target, \sys{}, the no-drafts server and SGLang-Omni all hold up to 128 streams. No system had a failed request.

The no-drafts server also turns epochs off, so this gap mixes two effects. On utterances up to 10 s, which are never cut into epochs, \sys{}'s median TTFS at 1 to 64 streams is 20--26 ms, against 58--95 ms without drafts. Drafts help only while checkpoints finish before the commit. At 64 streams, at least 98\% of finals are checked against a draft, but at 128 streams only 21--29\% are. Letting 16 or 32 streams run checkpoints at once keeps 95\% or more checked, and on Pipecat-STT cuts the median TTFS at 128 streams from 138.5 ms to 38.2 and 32.6 ms.

\subsection{Other recognizers under load}
\label{sec:eval-other-load}

We also batched the other recognizers with finals before checkpoints and the same client, using client-side timing and the GPUs listed in \cref{tab:other-load}. Only Whisper large-v3-turbo meets a 100 ms p99 target: up to 80 streams on Pipecat-STT and 32 on WildASR, against 32 and 8 without drafts. The other three miss it even at one stream.

Once the GPU is busy with finals, checkpoints rarely finish in time, and drafts stop helping. From 32 streams, VibeVoice-ASR's medians with and without drafts are within 8\% of each other. From 64 streams, Nemotron's median with drafts is up to 16\% higher. Without drafts, batching alone changes up to 6.1\% of finals. With drafts, 93.5--100\% of finals stay identical at every load, and WER stays within 0.3 points of the no-drafts arm. The exception is VibeVoice-ASR runs where only one arm drew a clip on which the model loops (\cref{app:models}).

\subsection{Endpointing}
\label{sec:endpointing}

Our client commits with the last chunk of speech. A deployed client instead commits after a silence timeout, so we appended 300 or 600 ms of silence at 1, 32 and 64 streams. Measured from the end of speech, \sys{}'s median stays 59--108 ms below the no-drafts server's, at equal WER. At 32 and 64 streams, 96.7--98.1\% of its finals are identical to offline decodes of the padded audio.

\section{Related Work}
\label{sec:related}

\textbf{Streaming ASR.} Local agreement~\cite{liu2020partial,machacek2023whisperstreaming} keeps the text that consecutive hypotheses agree on, and Qwen3-ASR's toolkit fixes a prefix~\cite{qwen3asrtoolkit}. None of them checks earlier output against the model on the complete audio. Two-pass systems~\cite{sainath2019twopass} re-decode a streamed first pass with full context, but do not reuse it to reach an exact final sooner.

\textbf{Drafts.} Speculative decoding preserves the target model's output rule~\cite{leviathan2023specdec}; SpecASR uses a separate draft recognizer to accelerate offline ASR~\cite{specasr2025}. SSBD reuses previous translations as drafts for growing text input, with biased verification and a zero-bias variant~\cite{ssbd2025}. Speculative Listening drafts under assumed silence and verifies against arriving audio in already chunk-interleaved full-duplex speech models~\cite{specslisten2026}. \sys{} addresses streaming conversion of offline ASR: the audio, prompt and state invariants for full-utterance greedy finals, adaptations across four model families, and serving measurements under load.

\section{Limitations and Future Work}
\label{sec:limits}

Our benchmarks are English, and Pipecat-STT's references are machine-generated. Batch-dependent GPU rounding changes some finals; batch-invariant kernels~\cite{he2025nondeterminism} could reduce this. \sys{} requires the model properties in \cref{sec:design}. Checkpoints consume GPU time and help slower models mainly at low load. The scheduler favors finals; improving partial latency remains future work.

\section{Conclusion}

\sys{} turns partial transcripts into verified drafts for offline ASR. Under the stated invariants, finals equal greedy offline output in exact arithmetic. The five-recognizer comparison retains 95--100\% byte-identical transcripts, and Qwen3-ASR meets a 100 ms p99 TTFS target at 64--80 concurrent streams on one A100.

\paragraph{Acknowledgments.} AI tools assisted with writing, literature search, implementation, and research analysis.

{\footnotesize\setlength{\bibsep}{1pt}
\bibliographystyle{abbrvnat}
\bibliography{references}}

\clearpage
\appendix

\section{Exactness and settings}
\label{app:exact}
\label{app:results}
\label{app:drafts}
\label{app:config}
\label{app:details}

\begin{table}[!h]
\centering\footnotesize
\caption{\textbf{Qwen3-ASR exactness by load.} Finals byte-identical to the offline reference (\%), and WER (\%) over each run's requests. N: no drafts. D: \sys{}.}
\label{tab:exact}
\setlength{\tabcolsep}{2.5pt}
\begin{tabular}{@{}rcccccccc@{}}
\toprule
& \multicolumn{4}{c}{Pipecat-STT} & \multicolumn{4}{c}{WildASR} \\
\cmidrule(lr){2-5}\cmidrule(l){6-9}
& \multicolumn{2}{c}{identical} & \multicolumn{2}{c}{WER} & \multicolumn{2}{c}{identical} & \multicolumn{2}{c}{WER} \\
$C$ & N & D & N & D & N & D & N & D \\
\midrule
\input{data-cap/exact-rows.tex} 
\bottomrule
\end{tabular}
\end{table}

\begin{table}[!h]
\centering\footnotesize
\caption{\textbf{Qwen3-ASR streams per A100 at a p99 TTFS target.} \sys{} and no drafts ran at the loads in \cref{tab:exact}; other systems at 1, 8, 32, 64 and 128 streams. Capacity is the largest tested load meeting the target at that and every lower load; ``--'' means not met even at one stream.}
\label{tab:capacity}
\setlength{\tabcolsep}{3pt}
\begin{tabular}{@{}lcccccc@{}}
\toprule
& \multicolumn{3}{c}{Pipecat-STT} & \multicolumn{3}{c}{WildASR} \\
\cmidrule(lr){2-4}\cmidrule(l){5-7}
p99 TTFS target & 100 ms & 200 ms & 300 ms & 100 ms & 200 ms & 300 ms \\
\midrule
\input{data-cap/capacity-both-rows.tex} 
\bottomrule
\end{tabular}
\end{table}

\begin{table}[!h]
\centering\footnotesize
\caption{\textbf{Capacity of the other recognizers.} Cells are streams served with no drafts $\to$ \sys{}, or one value when equal; ``--'' as in \cref{tab:capacity}; loads as for \sys{} there, with 16 and 48 also tested for VibeVoice-ASR. GPUs: A100-SXM4-40GB (Whisper), A100-SXM4-80GB (VibeVoice-ASR), RTX PRO 6000 96 GB (Nemotron).}
\label{tab:other-load}
\begin{tabular}{@{}lcccccc@{}}
\toprule
& \multicolumn{3}{c}{Pipecat-STT} & \multicolumn{3}{c}{WildASR} \\
\cmidrule(lr){2-4}\cmidrule(l){5-7}
p99 TTFS target & 100 ms & 300 ms & 1000 ms & 100 ms & 300 ms & 1000 ms \\
\midrule
\input{data-cap/other-load-rows.tex} 
\bottomrule
\end{tabular}
\end{table}

\textbf{Exactness on Qwen3-ASR.} \Cref{tab:exact} separates batching effects from drafts: without drafts, all finals match at one and eight streams, falling to 98.1--98.8\% at 128 streams. Drafts retain 96.8--99.3\% across loads. At one stream, epochs cut 42\% of Pipecat-STT and 11\% of WildASR utterances; 97.9\% and 100\% of those finals still match, respectively.

\textbf{Qwen3-ASR baselines.} Cross-request caches are disabled because the client replays utterances. SGLang-Omni and vLLM score 1.75\% and 1.77\% WER on Pipecat-STT; both score 4.70\% on WildASR.

\textbf{Other models.}
\label{app:models}
\Cref{tab:models} uses single-stream cloud A100 runs: 300 utterances per dataset for Whisper, all Pipecat-STT and 1,000 WildASR clips for VibeVoice-ASR and Nemotron. One WildASR clip makes every VibeVoice-ASR arm loop to its output limit, including offline, and is excluded. Whisper-Streaming uses faster-whisper defaults: whole-utterance decoding with those settings is within 2 WER points of its streaming result, while greedy decoding on that backend is within 0.2 points of our engine. Thus most of its gap comes from decoding settings. Batched runs use each engine's own offline reference on the same GPU and draw from all 2,549 WildASR clips. VibeVoice-ASR loops on two clips: excluding them gives 11.3--12.8\% WER across runs; including them raises it to as much as 20.2\%.

\clearpage
\begin{algorithm}[H]
\caption{\sys{} for one stream (Qwen3-ASR). $K$: the stream's KV cache. $d$: the latest draft. Continuation processes the last returned token unless it is end-of-sequence (EOS).}
\label{alg:devl}
\footnotesize
\begin{algorithmic}[1]
\Statex \textbf{initialize:} $K \gets$ prefill of the prompt prefix; $d \gets \emptyset$
\Function{Check}{$K$, new input, $d$} \Comment{one prefill pass per check}
  \State $k \gets |d|$; prefill the new input and $d_{1:\max(k-1,0)}$ on top of $K$
  \State let $\gamma_{1:\max(k,1)}$ be the next-token argmaxes at the draft positions
  \State $a \gets \max\{r \in \{0,\ldots,k\}: d_j=\gamma_j\ \text{for all }j\le r\}$
  \State $p \gets d$ if $a=k>0$, otherwise $d_{1:a}\,\gamma_{a+1}$ \Comment{empty draft uses $\gamma_1$}
  \State truncate $K$ after $p_{1:|p|-1}$ \Comment{keep the last token as the decode seed}
  \State if $a<k$ and $p$ does not end in EOS, realign the rest of $d$ after $p$ and extend $p$ by the same check, at most three repair rounds
  \State \Return $p$ \Comment{each repair preserves the cache convention above}
\EndFunction
\Statex \textbf{on new audio:}
\If{the input normalization changed} discard cached audio; reset $K$ to the prompt prefix
\EndIf
\State encode and prefill every complete block not currently cached \Comment{rebuild invalidated blocks}
\If{a checkpoint is due and the scheduler admits it}
  \State $p \gets$ \Call{Check}{copy of $K$, uncached audio and prompt suffix, $d$}
  \State greedily continue $p$; $d \gets$ the result; send $d$ as a partial; record the audio it saw
  \If{the epoch exceeds 10 s and $d$ ends at a stable sentence end} \Comment{epoch}
    \State emit that text; start a new epoch from shortly before the cut
  \EndIf
\EndIf
\Statex \textbf{on commit:}
\If{the last checkpoint saw all the audio and ended with end-of-sequence} \Return $d$ \Comment{promotion}
\EndIf
\State $p \gets$ \Call{Check}{$K$, tail and prompt suffix, $d$}; decode greedily after $p$ until end-of-sequence
\State \Return the result \Comment{residual decoding}
\end{algorithmic}
\end{algorithm}

\textbf{Correctness sketch.} Consider an utterance without epoch cuts, in exact arithmetic, with audio $x$ and greedy output $\ystar=\G(x)$.
\emph{Inputs.} A block is featurized only after its analysis windows are complete, with frames indexed from the stream's start. Rebuilding cached blocks when normalization changes makes their features equal to the offline features $F(x)$. Independent encoder windows and identical absolute token positions then make the cached encoder outputs and KV state equal to an offline prefill.
\emph{Verification.} If draft tokens before position $i$ match the greedy output, the input at $i$ is the offline prompt followed by $\ystar_{<i}$, so its argmax $\gamma_i=\ystar_i$. By induction, the accepted prefix and first correction form a prefix of $\ystar$. If all $k>0$ draft tokens match, the draft itself is such a prefix; no $\gamma_{k+1}$ is needed. An empty draft yields $\gamma_1=\ystar_1$. Truncating the cache before the last returned token and processing that token resumes the offline decode; EOS ends it. Repair rounds use the same test.
\emph{Promotion.} A completed checkpoint with the final sample count and normalization has the same offline prompt, so its EOS-terminated output is $\G(x)$.
\emph{Other models.} Whisper and Nemotron recompute the complete audio at the commit. Nemotron replays accepted tokens to restore the offline recurrent state; VibeVoice-ASR uses the deterministic front end described in \cref{sec:design}.

\end{document}

%% file: data-cap/exact-rows.tex
1 & 100.0 & 97.3 & 1.97 & 1.89 & 100.0 & 99.3 & 3.59 & 3.58 \\
8 & 100.0 & 96.8 & 1.80 & 1.77 & 100.0 & 98.5 & 3.92 & 4.00 \\
32 & 99.5 & 97.6 & 1.81 & 1.81 & 99.8 & 98.5 & 4.54 & 4.56 \\
64 & 99.3 & 98.0 & 1.77 & 1.77 & 99.5 & 98.5 & 4.65 & 4.65 \\
80 & 99.2 & 98.2 & 1.78 & 1.78 & 99.2 & 98.5 & 4.68 & 4.72 \\
96 & 98.8 & 98.1 & 1.83 & 1.81 & 99.1 & 98.7 & 4.79 & 4.81 \\
112 & 98.5 & 98.4 & 1.81 & 1.83 & 99.0 & 98.5 & 4.79 & 4.78 \\
128 & 98.1 & 98.3 & 1.77 & 1.77 & 98.8 & 98.8 & 4.81 & 4.82 \\

%% file: data-cap/capacity-both-rows.tex
no drafts & -- & 1 & 128 & -- & 32 & 128 \\
\sys{} & 80 & 80 & 128 & 64 & 96 & 128 \\
official streaming & -- & 32 & 64 & -- & 32 & 64 \\
SGLang-Omni & -- & 1 & 128 & -- & 1 & 128 \\
vLLM & -- & -- & 32 & -- & -- & 64 \\

%% file: data-cap/other-load-rows.tex
Whisper large-v3-turbo & 32$\to$80 & 128 & 128 & 8$\to$32 & 128 & 128 \\
Whisper large-v3 & -- & --$\to$64 & 128 & -- & 8$\to$32 & 128 \\
VibeVoice-ASR & -- & -- & --$\to$8 & -- & -- & 1 \\
Nemotron-3.5-Transcribe & -- & --$\to$8 & 8$\to$32 & -- & --$\to$1 & 8$\to$32 \\